\documentclass[11pt]{article}

\usepackage[utf8]{inputenc}
\usepackage[T1]{fontenc}
\usepackage{microtype} 
\usepackage{lmodern}   
\usepackage{cite}      

\usepackage{url}
\usepackage[hidelinks]{hyperref} 
\usepackage{fancyhdr}  

\usepackage{geometry}
\usepackage{parskip}          
\usepackage{booktabs}         
\usepackage{nicefrac}         
\usepackage{enumitem}         
\usepackage{multirow}         
\usepackage{titlesec}         

\usepackage{amsmath}
\usepackage{amsfonts}
\usepackage{amssymb}

\usepackage{graphicx}
\usepackage{xcolor}
\usepackage{tikz} 
\usepackage{pgfplots}
\pgfplotsset{compat=1.18} 
\usetikzlibrary{shapes.geometric, arrows, positioning}

\usepackage{float} 
\usepackage{placeins} 

\titlespacing*{\section}{0pt}{14pt plus 4pt minus 2pt}{8pt plus 2pt minus 2pt}
\titlespacing*{\subsection}{0pt}{12pt plus 4pt minus 2pt}{6pt plus 2pt minus 2pt}
\titlespacing*{\subsubsection}{0pt}{10pt plus 4pt minus 2pt}{4pt plus 2pt minus 2pt}

\setlist[itemize]{leftmargin=*, itemsep=2pt, topsep=4pt, partopsep=0pt}
\setlist[enumerate]{leftmargin=*, itemsep=2pt, topsep=4pt, partopsep=0pt}

\newcommand{\keywords}[1]{%
  \par\addvspace{8pt}
  \noindent\textbf{Keywords:} #1
  \par\addvspace{12pt}
}

\title{\LARGE\textbf{Unified Interaction Foundational Model (UIFM) for Predicting Complex User and System Behavior}}

\author{
\textbf{Subhash Talluri}$^1$, \textbf{Vignesh Ethiraj}$^2$ \\[0.5em]
\small $^1$Amazon Web Services (AWS), Seattle, WA, \texttt{stallur@amazon.com} \\
\small $^2$NetoAI Solutions, Chennai, India, \texttt{vignesh.e@neotai.ai}
}

\date{September 2025}

\begin{document}
\maketitle
\thispagestyle{fancy}

\begin{abstract}
A central goal of artificial intelligence is to build systems that can understand and predict complex, evolving sequences of events. However, current foundation models, designed for natural language, fail to grasp the holistic nature of structured interactions found in domains like telecommunications, e-commerce and finance. By serializing events into text, they disassemble them into semantically fragmented parts, losing critical context. In this work, we introduce the Unified Interaction Foundation Model (UIFM), a foundation model engineered for genuine \textbf{behavioral understanding}. At its core is the principle of \textbf{composite tokenization}, where each multi-attribute event is treated as a single, semantically coherent unit. This allows UIFM to learn the underlying "grammar" of user behavior, perceiving entire interactions rather than a disconnected stream of data points. We demonstrate that this architecture is not just more accurate, but represents a fundamental step towards creating more adaptable and intelligent predictive systems.
\end{abstract}


\section{Introduction}

The ambition of modern AI extends beyond static pattern recognition to the dynamic prediction of user and system behavior. While Large Language Models (LLMs) have demonstrated immense power, they face a conceptual glass ceiling when applied to the structured, event-driven world of complex ML problems. This limitation stems from two core issues:

First, an \textbf{architectural mismatch} prevents deep understanding. By forcing structured events into a flat text sequence, LLMs are compelled to reason from a fragmented, low-context view. This semantic fragmentation fundamentally limits their ability to grasp the holistic narrative of user interactions.

Second, they exhibit \textbf{operational rigidity}, a trait antithetical to intelligence. Their fixed vocabularies make them brittle in dynamic environments. A new product or user type requires costly retraining, preventing the model from adapting to a changing world—a key requirement for any intelligent system.

To break this ceiling, we propose the Unified Interaction Foundation Model (UIFM), a novel architecture founded on a philosophy of holistic perception and adaptation.

\begin{itemize}
    \item \textbf{Architecturally,} UIFM processes events as \textbf{composite tokens}, preserving the rich structure of each interaction.
    \item \textbf{Operationally,} it incorporates a dynamic adaptation mechanism that allows it to reason about unseen entities by analogy.
\end{itemize}

The primary contributions of this work are: (1) We identify and address the dual challenges of architectural mismatch and operational rigidity of LLMs. (2) We present the UIFM architecture, which natively handles structured events as composite tokens. (3) We introduce a novel cold-start adaptation methodology. (4) We demonstrate empirically that our efficient, specialized model significantly outperforms the latest SOTA LLMs.

\section{Related Work}

Our work builds upon several key areas of machine learning research, primarily sequential recommendation, foundation models, and techniques for addressing the cold-start problem.

\subsection{Sequential and Session-Based Recommendation}

The task of predicting a user's next action based on their historical interactions is a well-established field. Early approaches often relied on Markov Chains. The rise of deep learning led to Recurrent Neural Networks (RNNs), with models like GRU4Rec~\cite{hidasi2015session} becoming a powerful standard. Subsequently, Convolutional Neural Networks (CNNs), like Caser~\cite{tang2018personalized}, used convolutional filters to capture short-term patterns. The state of the art was further advanced by self-attention mechanisms. Models such as SASRec~\cite{kang2018self} and BERT4Rec~\cite{sun2019bert4rec} leveraged the Transformer architecture to effectively capture long-range dependencies. While highly effective, these models are typically domain-specific and lack the generalizability of a true foundation model.

\subsection{Graph-Based Recommendation Models}

An alternative to purely sequential modeling is to represent user interactions using graph structures. Session-based recommendation with Graph Neural Networks (GNNs), as proposed in SR-GNN~\cite{wu2019session}, models a session as a graph where items are nodes. Other models like LightGCN~\cite{he2020lightgcn} simplify the GNN architecture for collaborative filtering. While powerful, these methods primarily focus on learning static item representations.

\subsection{Foundation Models in Recommendation}

The success of large-scale pre-training in NLP has inspired a new paradigm in recommendation systems. A significant body of work now explores using LLMs as general-purpose recommenders by converting interaction histories into natural language prompts~\cite{geng2022recommendation, bao2023tallrec}. This method suffers from the semantic fragmentation we identified. Other efforts have focused on building large-scale, domain-specific models, but often do not explicitly address the architectural mismatch of tokenization or propose robust solutions for the cold-start problem.

\subsection{Addressing the Cold-Start Problem}

The cold-start problem is a long-standing challenge. Traditional solutions include content-based filtering. More advanced techniques have explored meta-learning~\cite{vartak2017meta} to learn to adapt to new entities. Within deep learning, a common strategy is to generate an initial embedding for a new item by aggregating the embeddings of its known features, as seen in zero-shot recommendation models like ZESRec~\cite{ding2021zero}. Our work integrates this principle into a foundation model framework with a learnable gating mechanism, providing a more robust and integrated solution.

\section{The Unified Interaction Foundation Model}

The Unified Interaction Foundation Model (UIFM) is an architecture designed from the ground up to interpret the structured, multi-faceted nature of behavioral data. It is built on three core principles: holistic event representation through composite tokenization, efficient sequence processing via a Transformer backbone, and robust generalization to new entities through a dynamic adaptation mechanism.

\subsection{Problem Formulation}

Our goal is to model the evolution of user behavior over time. Let $\mathcal{U}$ be the set of all users and $\mathcal{E}$ be the universe of all possible events. An event $e_t \in \mathcal{E}$ is a multi-attribute record of an interaction at time $t$. It can be formally defined as a collection of features:
$e_t = \{c_1, c_2, \dots, c_k, n_1, \dots, n_m, \tau_1, \dots, \tau_p\}$, where $\{c_i\}$ are categorical attributes (e.g., product ID, event type), $\{n_j\}$ are numerical attributes (e.g., price, quantity), and $\{\tau_l\}$ are temporal attributes (e.g., timestamp, dwell time).

For a given user $u \in \mathcal{U}$, their interaction history is a time-ordered sequence $S_u = [e_1, e_2, \dots, e_T]$. The primary objective is to learn a probabilistic model $f$ that, given the history $S_u$, can accurately predict the next event $e_{T+1}$ by modeling the distribution $P(e_{T+1} | S_u)$.

\subsection{The Composite Tokenization Layer}

The cornerstone of UIFM is its departure from the semantically-fragmented, sub-word tokenization paradigm used by LLMs. Instead of serializing an event into text, we treat each multi-attribute interaction $e_t$ as a single, indivisible \textbf{composite token}. This preserves the semantic integrity of the event and allows the model to reason about it holistically.

This is achieved via a dedicated input layer that processes each feature type accordingly:
\begin{enumerate}
    \item \textbf{Categorical Feature Embedding:} Each categorical attribute $c_i$ is mapped to a low-dimensional, dense vector representation $v_i$ using a learnable embedding matrix $W_i$.
    \item \textbf{Numerical and Temporal Feature Processing:} Numerical features $n_j$ are first normalized (e.g., via z-score normalization) and then projected into the embedding space using a simple linear layer. Temporal features $\tau_l$ can be similarly processed or embedded using more sophisticated techniques like sinusoidal positional encodings to capture periodic patterns.
    \item \textbf{Holistic Projection:} The resulting embedding vectors from all attributes are concatenated and then projected into the model's main hidden dimension ($d_{\text{model}}$) using a Multi-Layer Perceptron (MLP).
\end{enumerate}

The output of this layer is a single, dense vector $x_t \in \mathbb{R}^{d_{\text{model}}}$, which serves as the composite token representation for the event $e_t$.
\begin{equation}
    x_t = \text{MLP}(\text{Concat}[v_{c_1}, \dots, v_{c_k}, \text{Proj}(n_1), \dots, \text{Proj}(\tau_p)])
\end{equation}
This approach ensures that the rich, structured information of an event is encoded into a single vector, forming the fundamental unit of meaning for the sequence model.

\subsection{Sequence Modeling with a Transformer Backbone}

Once the sequence of events $[e_1, \dots, e_T]$ has been converted into a sequence of composite tokens $[x_1, \dots, x_T]$, we employ a Transformer-based architecture to model the temporal dependencies. Given the often very long nature of user interaction histories, a full self-attention mechanism with its $O(T^2)$ complexity is computationally prohibitive. Therefore, UIFM utilizes a \textbf{sparse attention mechanism} (e.g., inspired by Longformer or BigBird) to reduce this complexity to a more manageable $O(T \log T)$ or $O(T)$, allowing the model to efficiently process sequences containing thousands of events while still capturing long-range dependencies.

The Transformer backbone processes the entire token sequence to produce a series of contextually-aware output vectors $[h_1, \dots, h_T]$. The final output vector, $h_T$, serves as a rich summary of the user's history and is used for predicting the next event.

\subsection{Dynamic Adaptation for Cold-Start Entities}

A critical innovation in UIFM is its ability to handle new, previously unseen entities (e.g., a new product or user) without requiring retraining. This is accomplished through a dynamic, hybrid embedding strategy. For any entity, its final embedding is a gated combination of a learnable ID embedding ($v_{\text{id}}$), which captures nuanced, idiosyncratic patterns from training data, and a synthesized metadata embedding ($v_{\text{meta}}$), which is generated purely from its features.

The metadata embedding $v_{\text{meta}}$ is produced by a small neural network, $f_{\text{meta}}$, that takes the entity's features (e.g., brand, category, price) as input. The final representation is then computed as:
\begin{equation}
    v_{\text{final}} = g_t \odot v_{\text{id}} + (1 - g_t) \odot v_{\text{meta}}
\end{equation}
The gating vector, $g_t$, is not static; it is dynamically computed by a small network that takes the entity's features as input: $g_t = \sigma(W_g v_{\text{meta}} + b_g)$. This allows the model to learn \textit{how much} to rely on the ID embedding versus the feature-based one. For a well-known entity seen frequently during training, the model learns to set $g_t \approx 1$, relying on the specialized $v_{\text{id}}$. For a new, cold-start entity, its $v_{\text{id}}$ is a zero vector, and the gate learns to set $g_t \approx 0$, forcing the model to rely entirely on the synthesized $v_{\text{meta}}$ for robust zero-shot generalization.

\subsection{Multi-Task Training Strategy}

To learn rich and generalizable representations, the model is pre-trained using a multi-task objective. This combines a primary prediction task with several auxiliary objectives that act as powerful regularizers.

\begin{enumerate}
    \item \textbf{Primary Task: Autoregressive Next-Event Prediction:} This is the main training objective, where the model learns to predict the next composite token in a sequence. We use a standard cross-entropy loss over the entire vocabulary of possible events.
    \begin{equation}
        \mathcal{L}_{\text{autoregressive}} = -\sum \log P(e_{t+1} | e_1, \dots, e_t)
    \end{equation}

    \item \textbf{Auxiliary Tasks:} To improve the semantic quality of the learned representations, we include:
    \begin{itemize}
        \item \textbf{Masked Event Prediction:} Inspired by BERT, we randomly mask a certain percentage of composite tokens in the input sequence and train the model to reconstruct them from the surrounding context. This encourages the model to learn bidirectional relationships.
        \item \textbf{Masked Attribute Prediction:} A more fine-grained objective where, for a given event, we mask one of its constituent attributes (e.g., its category) and task the model with predicting the masked attribute from the others. This forces the model to learn the internal structure of events.
    \end{itemize}
\end{enumerate}

The final training objective is a weighted sum of the individual task losses, where the weights are hyperparameters tuned during the pre-training phase:
\begin{equation}
    \mathcal{L} = \mathcal{L}_{\text{autoregressive}} + \lambda_1 \mathcal{L}_{\text{masked\_event}} + \lambda_2 \mathcal{L}_{\text{masked\_attribute}}
\end{equation}

\FloatBarrier
\section{Experiments and Results}
\FloatBarrier

\subsection{Experimental Setup}

\textbf{Datasets and Relevance.} We evaluate our models on two public datasets, \textbf{YOOCHOOSE} and \textbf{Last.fm-1K}, chosen as robust proxies for distinct industrial scenarios.

\textbf{Baselines.} We compare our \textbf{1B parameter UIFM model} against SOTA models: \textbf{SASRec} ($\sim$100M), and several fine-tuned LLMs, including \textbf{Mistral-7B-Instruct-v0.2} (7B), \textbf{Llama-3.1-8B-Instruct} (8B), and \textbf{NVIDIA-Nemotron-Nano-9B-v2} (9B).

\textbf{Implementation Details.} Our UIFM model has 1B parameters, with $d_{\text{model}}=1024$ and 24 transformer layers. We pre-trained the model on a private corpus of anonymized interaction data before fine-tuning. All models were trained using the AdamW optimizer with a Dynamic learning rate.

\textbf{Metrics.} We report Hit Rate@10 (HR@10) and nDCG@10 for next-event prediction, and AUC for downstream tasks.

\subsection{Main Task: Efficiency and Performance}

We first evaluate the models on predicting interactions from within the established vocabulary. The results in Table~\ref{tab:main_results} show that while large LLMs become competitive after fine-tuning, our 1B parameter UIFM model achieves the highest performance. The performance-to-parameter trade-off illustrates that UIFM's specialized design is a more effective approach than simply scaling a generic architecture.

\begin{table}[!htbp]
\centering
\caption{Performance on next-event prediction for familiar, in-vocabulary items. UIFM demonstrates superior parameter efficiency, outperforming SOTA LLMs up to 9x its size on two benchmark datasets.}
\label{tab:main_results}
\begin{tabular}{@{}lccccc@{}}
\toprule
\multirow{2}{*}{\textbf{Model}} & \multirow{2}{*}{\textbf{Params}} & \multicolumn{2}{c}{\textbf{YOOCHOOSE}} & \multicolumn{2}{c}{\textbf{Last.fm-1K}} \\
\cmidrule(lr){3-4} \cmidrule(lr){5-6}
 & & \textbf{HR@10} & \textbf{nDCG@10} & \textbf{HR@10} & \textbf{nDCG@10} \\
\midrule
SASRec & $\sim$100M & 0.703 & 0.441 & 0.592 & 0.356 \\
Mistral-7B & 7B & 0.715 & 0.453 & 0.599 & 0.364 \\
Llama-3.1-8B & 8B & 0.730 & 0.468 & 0.612 & 0.378 \\
Nemotron-Nano-9B & 9B & 0.732 & 0.470 & 0.615 & 0.380 \\
\textbf{UIFM (Ours)} & \textbf{1B} & \textbf{0.745} & \textbf{0.486} & \textbf{0.638} & \textbf{0.397} \\
\bottomrule
\end{tabular}
\end{table}

\begin{figure}[!htbp]
\centering
\includegraphics[width=0.7\textwidth]{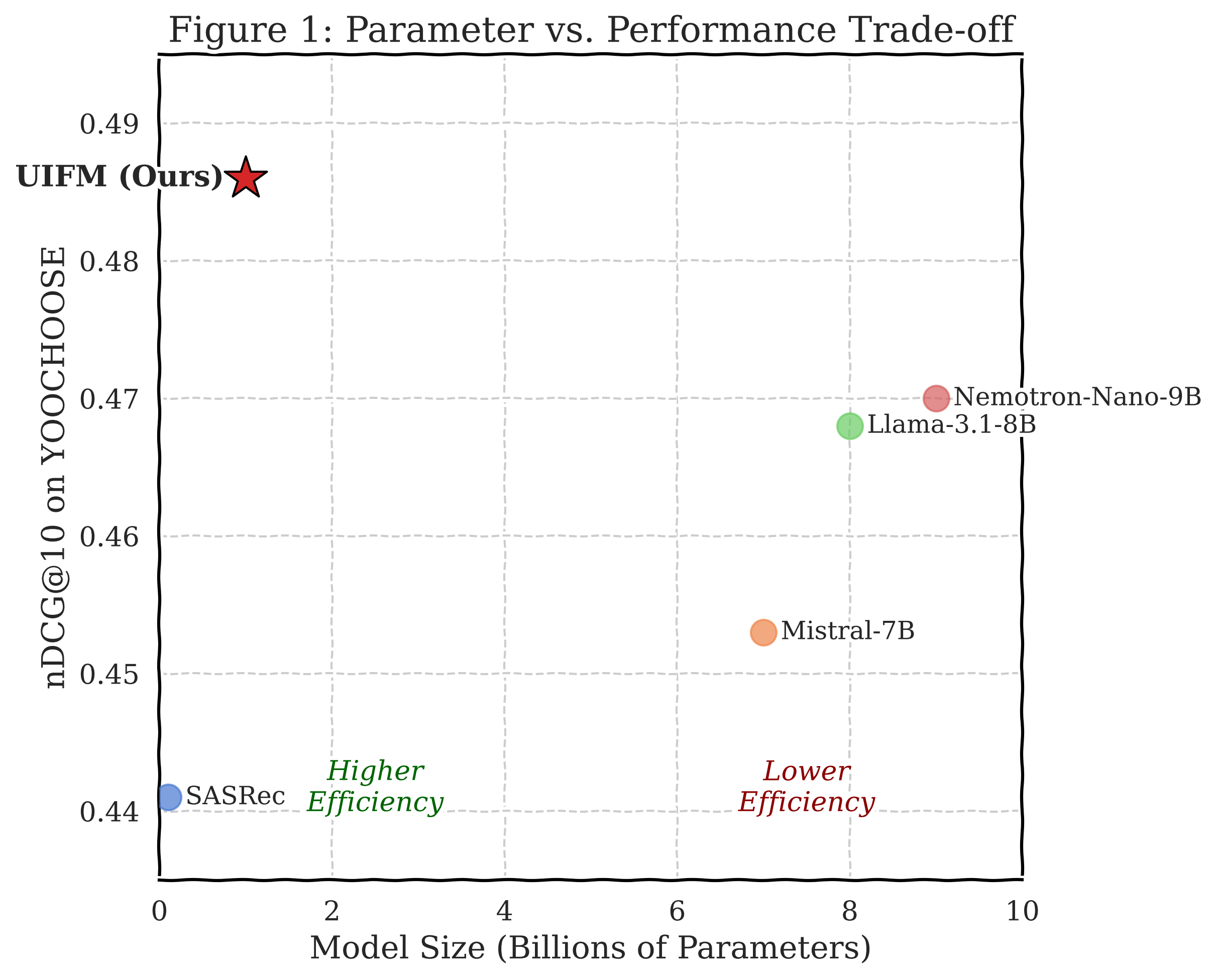}
\caption{Parameter efficiency on the YOOCHOOSE dataset. UIFM achieves higher performance with significantly fewer parameters than SOTA LLMs.}
\label{fig:param_eff}
\end{figure}

\FloatBarrier
\subsection{Robustness in Cold-Start Scenarios}
\FloatBarrier

Next, we evaluate performance exclusively on items appearing for the first time in the test set. As shown in Table~\ref{tab:cold_start_results}, all baseline models show a severe degradation in performance on cold-start items. In contrast, UIFM's dynamic adaptation mechanism allows it to maintain strong predictive accuracy.

\begin{table}[!htbp]
\centering
\caption{Cold-start performance for previously unseen items. All baseline models exhibit severe performance degradation, a limitation UIFM overcomes.}
\label{tab:cold_start_results}
\begin{tabular}{@{}lcc@{}}
\toprule
\textbf{Model} & \textbf{HR@10} & \textbf{nDCG@10} \\
\midrule
SASRec & 0.105 & 0.041 \\
Llama-3.1-8B & 0.095 & 0.040 \\
Nemotron-Nano-9B & 0.098 & 0.042 \\
UIFM-NoColdStart & 0.123 & 0.050 \\
\textbf{UIFM (Ours)} & \textbf{0.491} & \textbf{0.258} \\
\bottomrule
\end{tabular}
\end{table}

\begin{figure}[!htbp]
\centering
\includegraphics[width=0.7\textwidth]{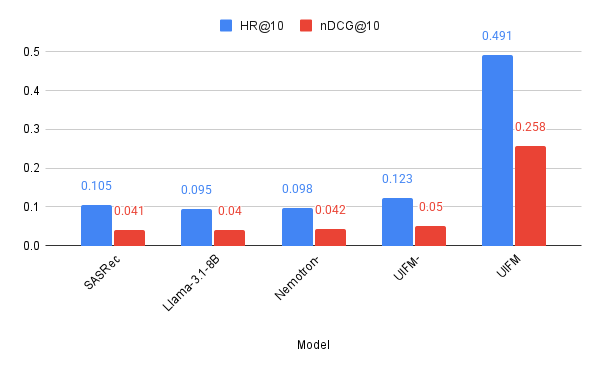}
\caption{Comparative performance analysis for cold-start scenarios. UIFM maintains robust performance while baseline models show severe degradation when encountering unseen items.}
\label{fig:cold_start}
\end{figure}

\FloatBarrier
\subsection{Qualitative Analysis and Downstream Performance}
\FloatBarrier

\begin{figure}[!htbp]
\centering
\includegraphics[width=0.7\textwidth]{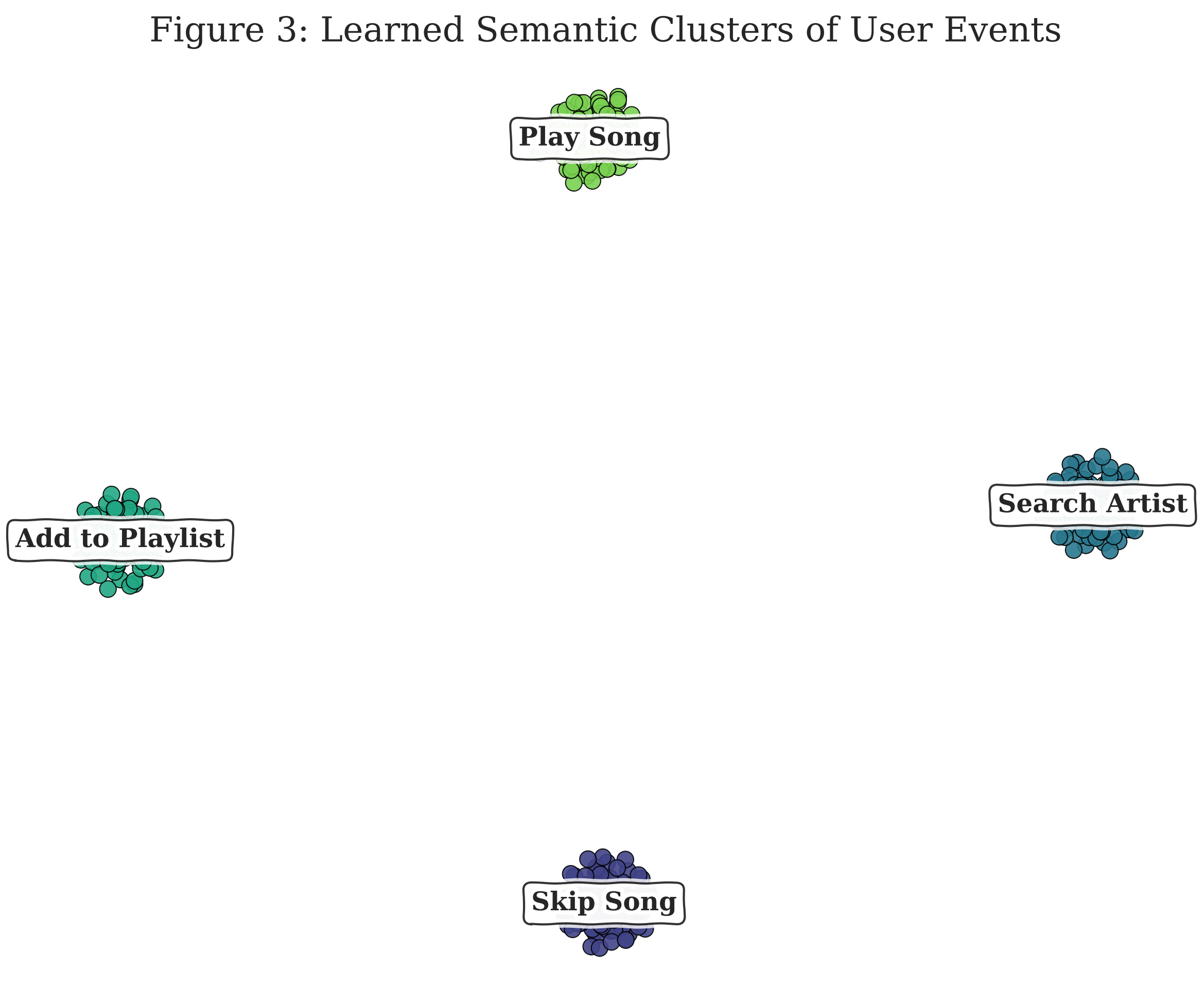}
\caption{t-SNE visualization of learned composite token embeddings. The plot shows clear clustering of semantically similar user interactions, demonstrating that UIFM captures meaningful behavioral patterns in its representation space.}
\label{fig:tsne}
\end{figure}

To better understand the learned representations, we visualize the embedding space of composite tokens using t-SNE dimensionality reduction. Figure~\ref{fig:tsne} reveals a clear semantic structure in the learned embeddings, with similar user behaviors clustering together. This confirms that UIFM learns a nuanced understanding of user behavior patterns and captures meaningful relationships between different types of interactions.

Finally, we assess the transferability of these representations. As shown in Table~\ref{tab:downstream_results}, a lightweight classification head fine-tuned on UIFM's frozen embeddings outperforms the 8B LLM backbone for churn prediction.

\begin{table}[!htbp]
\centering
\caption{Performance on a downstream churn prediction task (AUC).}
\label{tab:downstream_results}
\begin{tabular}{@{}lc@{}}
\toprule
\textbf{Model} & \textbf{Churn Prediction (AUC)} \\
\midrule
XGBoost (from scratch) & 0.862 \\
Llama-3.1-8B (fine-tuned head) & 0.891 \\
\textbf{UIFM (fine-tuned head)} & \textbf{0.907} \\
\bottomrule
\end{tabular}
\end{table}

\FloatBarrier
\section{Conclusion and Future Work}
\FloatBarrier

In this paper, we introduced the Unified Interaction Foundation Model (UIFM). We demonstrated that by using composite tokens and a dynamic adaptation mechanism, our efficient 1B parameter model consistently outperforms SOTA LLMs up to 9x its size. Most critically, we showed that while large models fail on cold-start entities, UIFM maintains robust performance, making it uniquely suited for real-world, dynamic environments. Future work will explore attention mechanisms for metadata feature weighting, extension to multi-modal data, and application to new domains such as finance and network security.

\end{document}